\documentclass[lettersize,journal]{IEEEtran}
\usepackage{amsmath,amsfonts}
\usepackage{algorithmic}
\usepackage{subfigure}
\usepackage{algorithm}
\usepackage{array}
\usepackage[caption=false,font=normalsize,labelfont=sf,textfont=sf]{subfig}
\usepackage{textcomp}
\usepackage{stfloats}
\usepackage{url}
\usepackage{verbatim}
\usepackage{booktabs}
\usepackage{graphicx}
\usepackage{cite}
\usepackage{caption}
\begin{document}

\title{SHA-CNN: Scalable Hierarchical Aware Convolutional Neural Network for Edge AI}

\author{Narendra Singh Dhakad,~\IEEEmembership{Graduate Student Member,~IEEE,} Yuvnish Malhotra, Santosh Kumar Vishvakarma,~\IEEEmembership{Senior Member,~IEEE,} Kaushik Roy,~\IEEEmembership{Fellow,~IEEE}

\thanks{Narendra Singh Dhakad, Yuvnish Malhotra and Santosh Kumar Vishvakarma are with Nanoscale
Devices, VLSI Circuit and System Design (NSDCS), Department of Electrical Engineering, Indian Institute of Technology Indore, MP 453552, India.

Kaushik Roy is with the School of Electrical and Computer Engineering, Purdue University, West Lafayette, IN 47907 USA.}
}

\maketitle

\begin{abstract}
This paper introduces a Scalable Hierarchical Aware Convolutional Neural Network (SHA-CNN) model architecture for Edge AI applications. The proposed hierarchical CNN model is meticulously crafted to strike a balance between computational efficiency and accuracy, addressing the challenges posed by resource-constrained edge devices. SHA-CNN demonstrates its efficacy by achieving accuracy comparable to state-of-the-art hierarchical models while outperforming baseline models in accuracy metrics. The key innovation lies in the model's hierarchical awareness, enabling it to discern and prioritize relevant features at multiple levels of abstraction. The proposed architecture classifies data in a hierarchical manner, facilitating a nuanced understanding of complex features within the datasets. Moreover, SHA-CNN exhibits a remarkable capacity for scalability, allowing for the seamless incorporation of new classes. This flexibility is particularly advantageous in dynamic environments where the model needs to adapt to evolving datasets and accommodate additional classes without the need for extensive retraining. Testing has been conducted on the PYNQ Z2 FPGA board to validate the proposed model. The results achieved an accuracy of 99.34\%, 83.35\%, and 63.66\% for MNIST, CIFAR-10, and CIFAR-100 datasets, respectively. For CIFAR-100 our proposed architecture perform hierarchical classification with 10\% reduced computation while compromising only 0.7\% of accuracy with the state-of-the-art. The adaptability of SHA-CNN to FPGA architecture underscores its potential for deployment in edge devices, where computational resources are limited. The SHA-CNN framework thus emerges as a promising advancement in the intersection of hierarchical CNNs, scalability, and FPGA-based Edge AI.
\end{abstract}


\section{Introduction}
\IEEEPARstart {C}onvolutional Neural Networks (CNNs) have been widely used in computer vision applications such as image classification, object detection, and segmentation \cite{imagenet}. However, as the size and complexity of datasets continue to grow, there is a need for more scalable and efficient CNN architectures that can handle large amounts of data while maintaining high accuracy. One approach to addressing this challenge is to use hierarchical CNN architectures consisting of multiple feature extraction and abstraction levels. Hierarchical CNNs have been shown to be effective in capturing complex visual patterns and improving the generalization performance of CNNs. However, existing hierarchical CNN architectures \cite{BCNN, HDCNN, ConditionCNN, cfcnn, hierarchynet, learncnn, encoding, GHCNN, hierarchyCNN, DeeplyCNN, multimodelCNN, CNNRNN} often suffer from scalability issues, such as high computational and memory requirements, which limit their applicability to large-scale datasets. Therefore, there is a need for a scalable hierarchical CNN architecture that can handle large amounts of data while maintaining high accuracy. Similar work has been proposed \cite{treecnn}, where they addressed the catastrophic forgetting concept to learn the new classes.

\begin{figure}[t]
		\centering
		\subfigure[A dolphin and whale; both belong to the same supper class, “mammals”]{\includegraphics[scale=0.5]{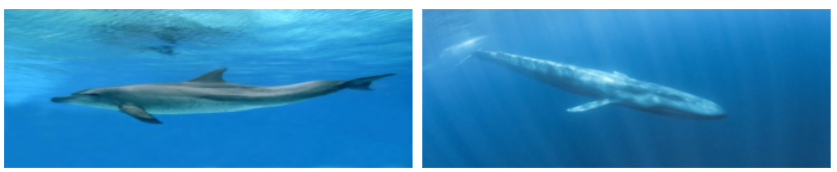}}
		\subfigure[Traffic sign in day and night]{\includegraphics[scale=0.5]{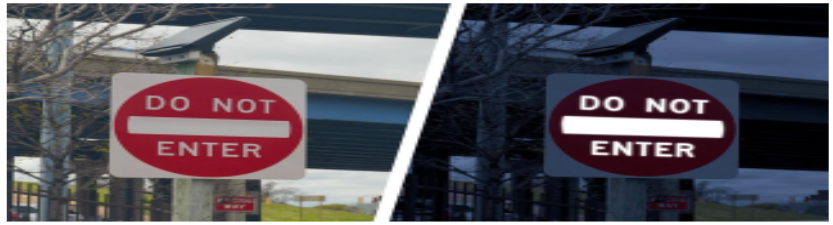}}
		\caption{Importance of hierarchical classification.}
\label{dolphine}
\end{figure}

\begin{figure*}[t]
    \centering
	\includegraphics[scale=0.45]{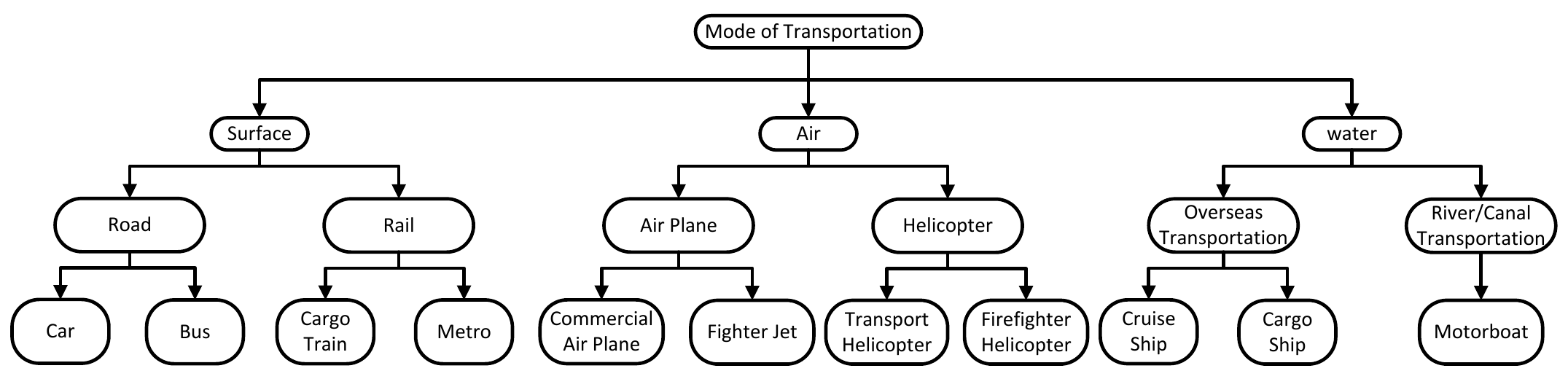}
	\caption{Example of a hierarchical tree for various classes of transportation modes.}
	\label{hierarchy}
\end{figure*}

Conventional CNNs are trained as a flat N-way classifier, meaning all the dataset classes are considered equally difficult to distinguish from each other and exclusive. However, some classes may be more difficult to distinguish than others as they may depend on higher hierarchy levels. For example, an apple and an orange belong to a common superclass of fruits and have some similarities. An apple and a zebra are easily distinguishable, but an apple and an orange are relatively difficult to distinguish. From a human perspective, if a human is given an image of an apple and an orange, the brain will first classify both as fruits and then the classification will go deeper to specific fruit for each image. More examples of such class dependencies from ImageNet dataset \cite{dataset} are shown in Fig. \ref{dolphine}. An example of a hierarchical tree based on class dependencies for different modes of transportation is shown in Fig. \ref{hierarchy}. Such an approach of classification following the tree structure going from superclasses to subclasses reduces the region of interest of classification in the final step, indeed improving the accuracy. NNs fail to encode such hierarchy or dependencies in classes into their architecture in their conventional form. This is where the concept of hierarchical classification comes into the picture. 

Scalability is a significant challenge in CNNs as the size and complexity of the input images and the number of parameters in the model can quickly become very large, making training and inference computationally expensive and memory-intensive. One main factor contributing to scalability issues in CNNs is the number of convolutional layers and filters in the model. Increasing the number of layers and filters can improve the model's accuracy but also increase the computational cost and memory requirements. This can limit the size of the input images or the batch size during training and make it challenging to deploy the model on resource-constrained devices or in real-time applications. For the N level of hierarchical classification, we need to add N branches, which improves the computational parameters linearly. 

We proposed a modified CNN model for scalable and hierarchical-aware CNN to address the issue of learning new classes and classification in a hierarchical approach. The motivation behind scalable hierarchical-aware CNN is to address the scalability challenge in CNNs by designing efficient, compact, and adaptable models to different input image sizes and resolutions. Traditional CNNs use fixed-size convolutional filters and feature maps, which can limit their ability to generalize to images with different sizes or aspect ratios. This can be especially problematic for object detection and semantic segmentation applications, where the input images can vary in size and aspect ratio. Scalable hierarchical-aware CNNs aim to overcome these limitations by incorporating multi-scale and hierarchical representations in the model architecture. This allows the model to process images at multiple resolutions and scales, capturing both fine-grained and coarse-grained visual patterns in the input images. 

The contribution of the paper is as follows: 
\begin{enumerate}
    \item Modified a hierarchical CNN architecture that can effectively capture complex visual patterns while minimizing computations, memory requirements and delay for inference.
    \item The proposed model architecture supports the scalability if the number of classes increases. 
    \item Evaluate the scalability and performance of the proposed approach on large-scale datasets and compare it with existing state-of-the-art methods.
    \item The proposed model architecture has been deployed to the PYNQ Z2 FPGA board and performance has been evaluated for MNIST, CIFAR-10, CIFAR-100 datasets.
\end{enumerate}

The rest of the paper is organized as follows. The related work on hierarchical and incremental learning is discussed in Section II. In Section III, we present our proposed network architecture. In Section IV, the experimental performance and hardware demonstration of the model. Finally, Section V concludes the work.


\section{Background and Related Work}
Conventional neural networks are considered an N-way classifier with independent final categories. However, a few classes may be more difficult to distinguish from one another than the others as they may depend on a higher level of hierarchy. There are numerous works have been proposed to address hierarchical learning \cite{BCNN, HDCNN, ConditionCNN, cfcnn, hierarchynet, learncnn, GHCNN, hierarchyCNN, DeeplyCNN, multimodelCNN, CNNRNN}, while only a few works address the issue of scalability \cite{treecnn}. 

B-CNN \cite{BCNN} is designed to leverage the hierarchical structure of the classification problem. It consists of two branches of convolutional neural networks: one for extracting fine-grained features and the other for extracting coarse-grained features. The fine-grained branch is designed to capture subtle details and local features, while the coarse-grained branch captures more general and global features. The two branches are combined to form a joint feature representation fed into a hierarchical classification classifier. B-CNN uses the fine-grained branch to refine the feature representation based on the subclass label and the coarse-grained branch to refine the feature representation based on the superclass label. This approach allows the network to learn a discriminative feature representation for fine-grained and coarse-grained categories, improving the accuracy of hierarchical classification tasks. However, the work bottleneck is that if the number of levels increases in the hierarchy, the computation cost increases linearly, which is inefficient for edge devices.

HD-CNN \cite{HDCNN} presents a hierarchical deep convolutional neural network architecture for large-scale visual recognition tasks. The proposed architecture consists of multiple levels of feature extraction, where each level refines the input data representation. This approach improves the accuracy of recognition tasks, allowing for more complex and discriminative features to be extracted.

The article \cite{encoding} suggests that encoding hierarchical information in neural networks can help deal with subpopulation shift. This problem arises when the training and test data come from different subpopulations. The work proposes a new neural network architecture called HiGCN (Hierarchical Graph Convolutional Network) that utilizes the hierarchical structure of the data to improve performance on subpopulation shift tasks. The HiGCN model uses a graph-based approach to represent the hierarchical structure of the data, with nodes representing different levels of abstraction in the hierarchy. They demonstrate the effectiveness of the hierarchical graph structure in capturing and utilizing the hierarchical information in the data. 

\begin{figure*}[t]
	\centering
	\includegraphics[scale=0.25]{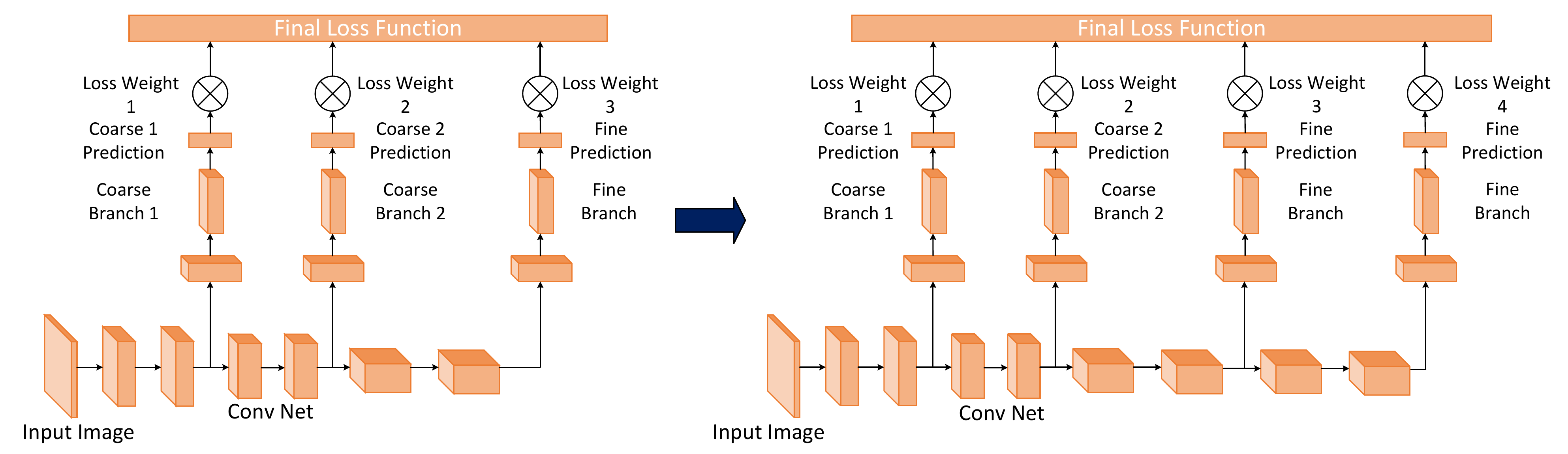}
	\caption{B-CNN model architecture \cite{BCNN}. The level increase resulted in the increase of another block and branch with a fully connected layer, leading to more parameters in the network.}
	\label{BCNN}
\end{figure*}

Condition-CNN \cite{ConditionCNN} proposed a hierarchical multi-label image classification model specifically designed for fashion images. It aims to classify images into multiple labels, such as the type of clothing, its style, color, and pattern. This utilizes a hierarchical approach, where the coarse label classifier acts as a gating mechanism for the fine label classifier. The fine label classifier is only activated if the coarse label classifier predicts the correct general category for the clothing item. This approach reduces the search space for the fine-label classifier, resulting in a more efficient and accurate classification process. 

CF-CNN \cite{cfcnn} proposed a coarse-to-fine convolutional neural network (CF-CNN) for learning multilabel classes, which is a hierarchical learning method that creates a class group with the hierarchical association and assigns a new label to each class belonging to the group so that each class acquires multiple labels. CF-CNN is a three-layer network with a main coarse layer, a main fine layer, and a refined layer. 

\begin{figure}[b]
	\centering
	\includegraphics[scale=0.25]{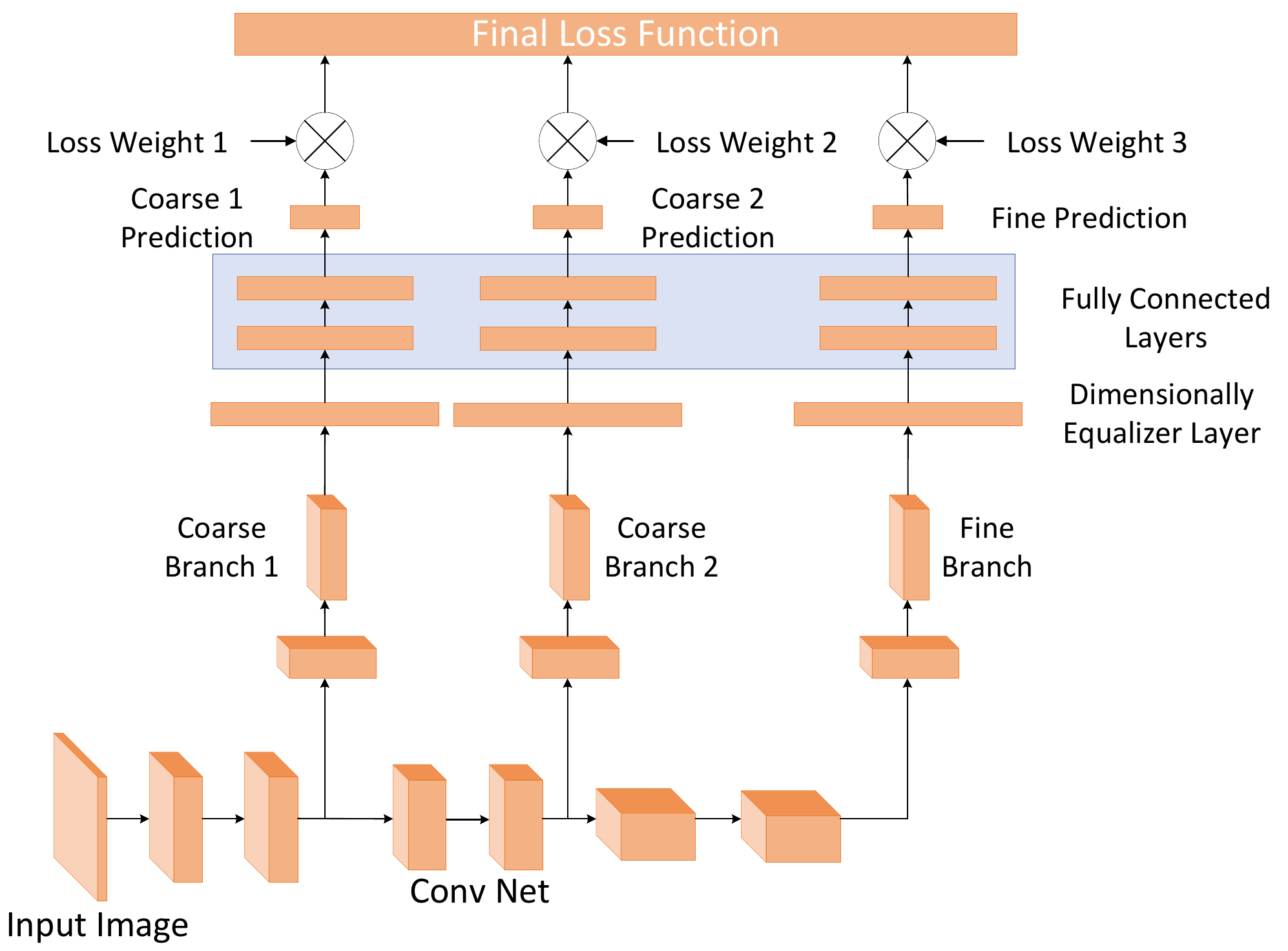}
	\caption{SHA-CNN: training and weight extraction.}
	\label{SHA1}
\end{figure}

Tree-CNN \cite{treecnn} explores the challenges of incorporating new data into fixed Deep CNNs due to catastrophic forgetting and sensitivity to hyperparameter tuning. The paper proposes an adaptive hierarchical network structure composed of DCNNs that can grow and learn as new data becomes available, organizing the incrementally available data into feature-driven superclasses. The paper compares this hierarchical model against fine-tuning a deep network and demonstrates a significant reduction of training effort while maintaining competitive accuracy. This classifier is composed of multiple nodes connected in a tree-like manner, where each node has a DCNN trained to classify the input into one of its children. The network adds branch nodes as new classes get grouped under the same output nodes. The work discusses the constraints applied to the algorithm, including the maximum depth of the tree and the maximum number of child nodes for a branch node. The hierarchical node-based learning model of the Tree-CNN attempts to confine the model change to only a few nodes, limiting the computation costs of retraining. 

The work in \cite{docnn?} discusses the impact of hierarchical structure on CNNs and how it affects their accuracy in image classification. They proposed a visual analytics method, which allows users to analyze the similarities between classes and the structure of the feature detectors learned by the CNNs in relation to the class hierarchy. The tool offers a hierarchy viewer and a seriation algorithm to reveal block patterns in a confusion matrix. Additionally, the tool helps diagnose various quality issues in the data and improve the curation of training datasets. They propose hierarchy-aware CNN architectures that yield significant gains in classification accuracy and convergence speed. 

An approach to hierarchical multi-label classification that uses a neural network to predict all categorizations of an input source, exploiting multiple layers of a neural model is presented in \cite{learncnn}. The approach is an architectural extension that can be adapted to a generic neural network using deep stack layers. The experiments demonstrate the ability of the hierarchical approach to recognize the correct categories of a class hierarchy and when a classification model fails to recognize the correct class. 

HierarchyNet \cite{hierarchynet} proposes an approach to a multi-label hierarchical classification of buildings that uses fewer parameters than existing hierarchical networks and achieves more accurate classification. The proposed model can extract features and classify buildings based on functional purpose and architectural style using a coarse-to-fine hierarchical structure with a custom multiplicative layer. The model is based on the B-CNN \cite{BCNN}, which is scalable for two-level hierarchical classification but not for more levels. 

\begin{figure*}[t]
	\centering
	\includegraphics[scale=0.4]{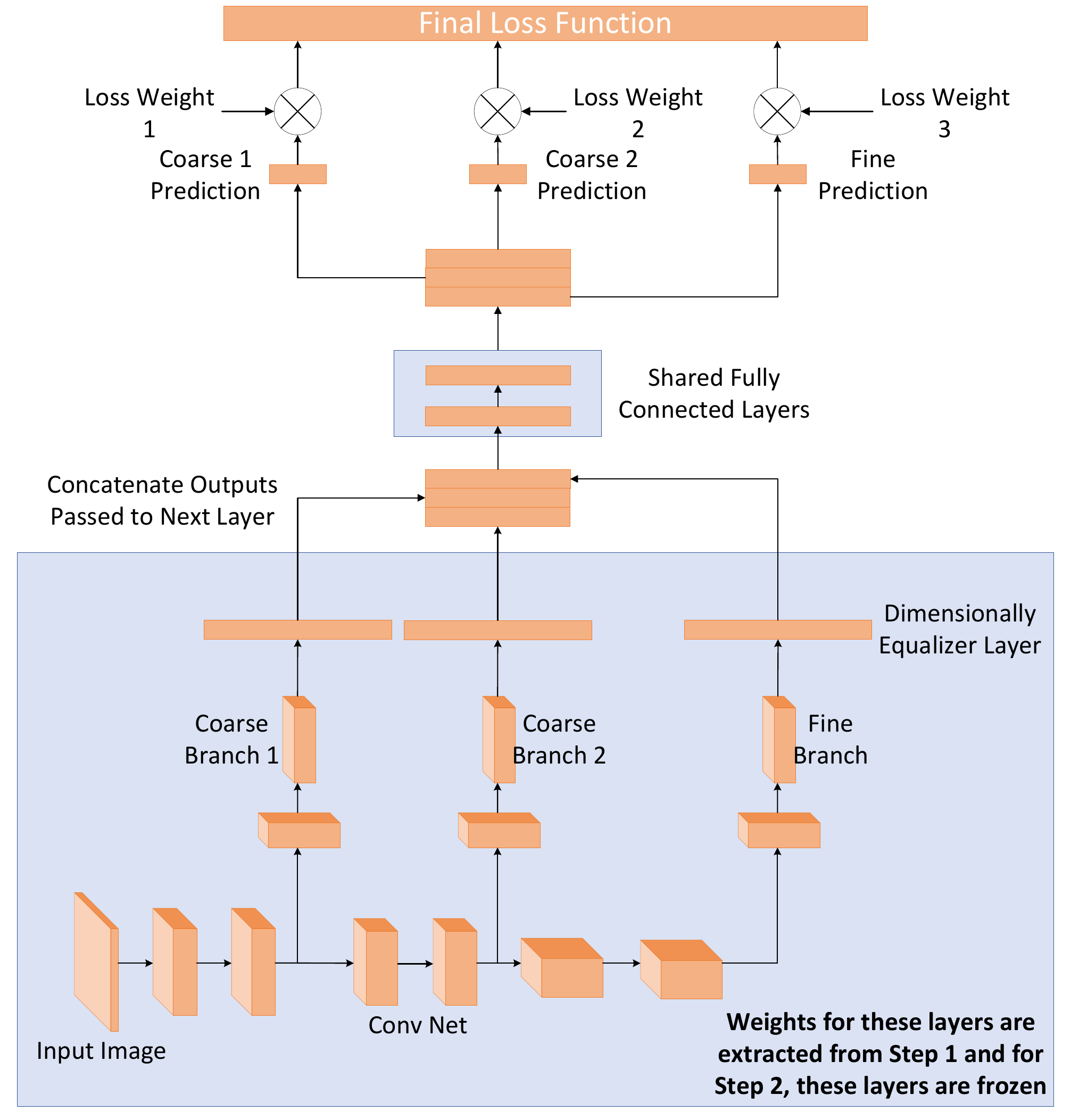}
	\caption{SHA-CNN: shared fully connected layers for all branches.}
	\label{SHA2}
\end{figure*}

\begin{figure*}[t]
	\centering
	\subfigure[MNIST dataset label tree]{\includegraphics[scale=0.45]{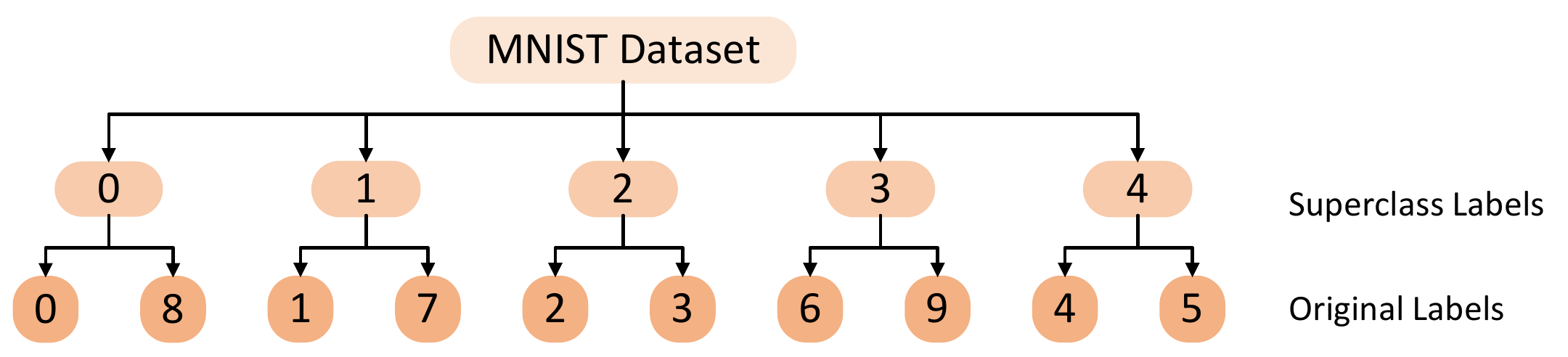}}
    \subfigure[CIFAR-10 dataset label tree]{\includegraphics[scale=0.47]{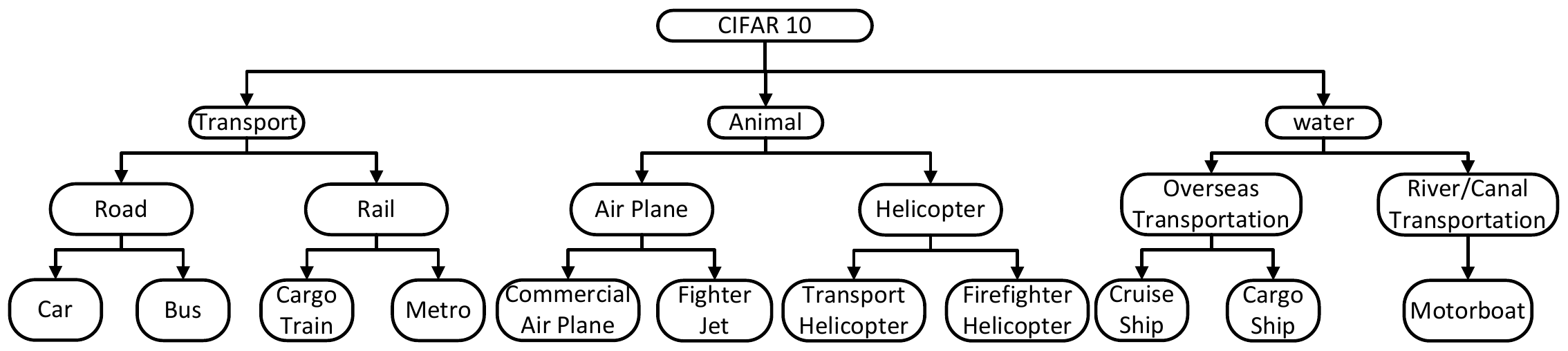}}
	\caption{Example of label tree classification for MNIST and CIFAR-10 datasets.}
	\label{MNISTCifar}
\end{figure*}

\begin{figure}[t]
	\centering
	\includegraphics[scale=0.4]{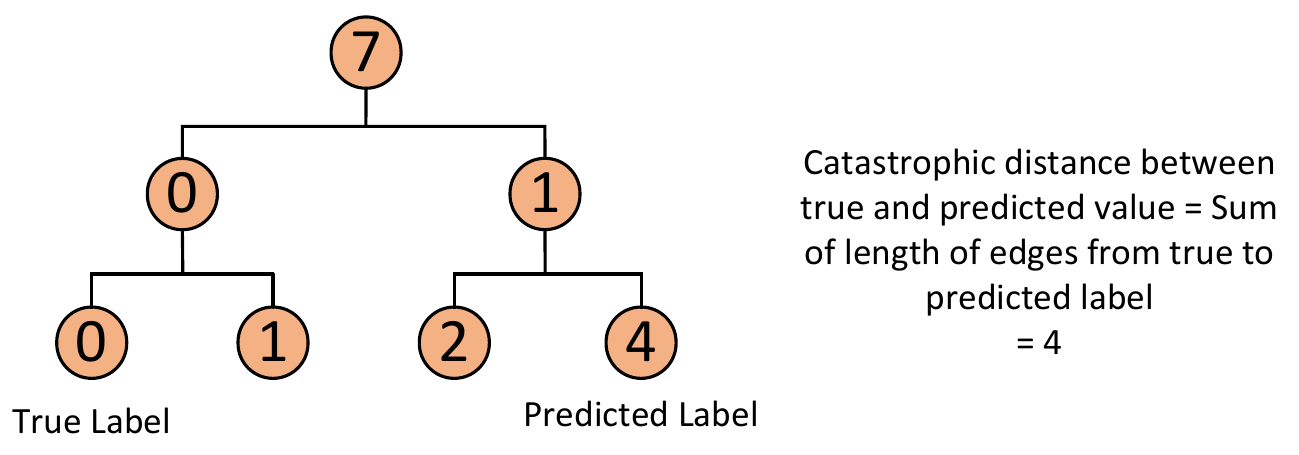}
	\caption{Example of catastrophic distance calculation in the proposed model}
	\label{cat_distance}
\end{figure}


\section{Proposed Model Description}
In our proposed work, we considered the model presented in B-CNN \cite{BCNN} as a reference. The model architecture B-CNN is built upon one important characteristic of CNNs and their architecture - lower layers of CNNs capture low-level features like edges and corners. In comparison, deeper layers extract higher-level features like faces, eyes etc. These deeper layers are built on top of lower levels, meaning higher-level features are extracted from lower-level features. This completely aligns with the hierarchical classification required. B-CNN model leverages this property of CNN by adding branches along the architecture flow, each branch making predictions at each level of the label tree. On top of each branch, fully connected layers and a softmax layer are used to produce the output in a one-hot encoding fashion. The branches may contain convolutional and pooling layers as well. However, in the experimentation of B-CNN models, branches only contained fully connected layers. As the number of levels increases in the label tree, the number of branches increases, increasing the number of computational parameters in the network. This makes scaling difficult, especially for edge devices with limited computational resources. Our proposed model architecture and training method will make the final model scalable at inference. The procedure includes two steps: (i) B-CNN training and trained weights extraction, and (ii) combining fully connected layers of each branch and using a single fully connected set for prediction at each level. The details of each step are mentioned in the following sub-sections. 

\subsection{B-CNN Training and Weights Extraction}
CNNs are made of two components: the feature extraction component, which consists of convolutional and pooling layers, and the classification component, which consists of fully connected layers for prediction based on features extracted by the feature extraction component. B-CNN models are trained using the Branch Training Strategy - the loss weights are changed after specific epochs so that the training focus stays at the specific branch for some epochs and then the focus changes to the next branch. This helps the Feature Extraction component of CNN to learn weights capable of extracting features specific to a specific level of the label tree. So in our procedure, firstly, the B-CNN model in trained on the dataset and the label tree. One change from the B-CNN model is adding a dense layer to each branch. This additional dense layer will have same number of hidden units in all the branches. Weights of convolutional layers and the additional dense layer of each branch are saved. These weights will be used in the second step. Model architecture for weight extraction is shown in Fig. \ref{SHA1}.

\subsection{Shared Fully Connected Layers at all levels}
Fully connected layers can be considered a mathematical function followed by an activation function mapping features to final predictions. The idea behind shared fully connected layers at each level prediction is to have a mathematical function with weights and parameters learned to predict at all levels of the label tree. The model architecture for this step is shown in Fig. 5. As can be seen, the softmax layer for each branch is kept separate, as there may be a different number of labels at each level of the label tree. While training the model, the layers for which the weights were extracted from the model from the previous step are kept frozen, meaning the weights of these layers are not trained in this step. Only the shared fully connected layers and the softmax layer of each branch are trained at a low learning rate. This model will be used at inference once trained. Having shared fully connected layers in the network will reduce the number of parameters in the model at inference which in turn will make the model more scalable but also, less storage space will be required for storing the model weights.

\begin{figure}[t]
	\centering
	\includegraphics[scale=0.4]{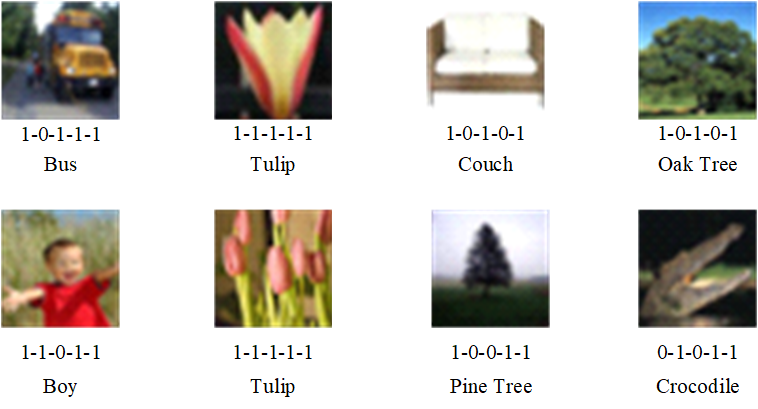}
	\caption{Sample images with prediction at different levels.}
	\label{images}
\end{figure}

\begin{table*}[t]
\centering
\caption{\scshape Accuracy and catastrophic distance comparison of proposed model with state of the art}\label{comparison}
\renewcommand{\arraystretch}{1.31}
\resizebox{\linewidth}{!}{
\begin{tabular}{l|cc|cc|cc} \toprule
& \multicolumn{2}{c|}{\textbf{Baseline VGG16}}   & \multicolumn{2}{c|}{\textbf{B-CNN \cite{BCNN}}}   & \multicolumn{2}{c}{\textbf{SHA-CNN}}\\ \midrule
\multicolumn{1}{c|}{\textbf{Dataset}} & \multicolumn{1}{c}{\textbf{Accuracy (\%)}} & \multicolumn{1}{c|}{\textbf{Catastrophic Distance}} & \multicolumn{1}{c}{\textbf{Accuracy (\%)}} & \multicolumn{1}{c|}{\textbf{Catastrophic Distance}} & \multicolumn{1}{c}{\textbf{Accuracy (\%)}} & \multicolumn{1}{c}{\textbf{Catastrophic Distance}} \\ \midrule
\textbf{MNIST}    & 99.44      & 0.01      & 99.39       & 0.01      & 99.34    & 0.015     \\ 
\textbf{CIFAR-10} & 82.56      & 0.68      & 84.02       & 0.607     & 83.35    & 0.6328    \\ 
\textbf{CIFAR-100}& 62.89      & 1.6576    & 64.14       & 1.569     & 63.66    & 1.5968    \\
\bottomrule
\end{tabular}}
\end{table*}

\begin{figure}[t]
	\centering
	\includegraphics[scale=0.06]{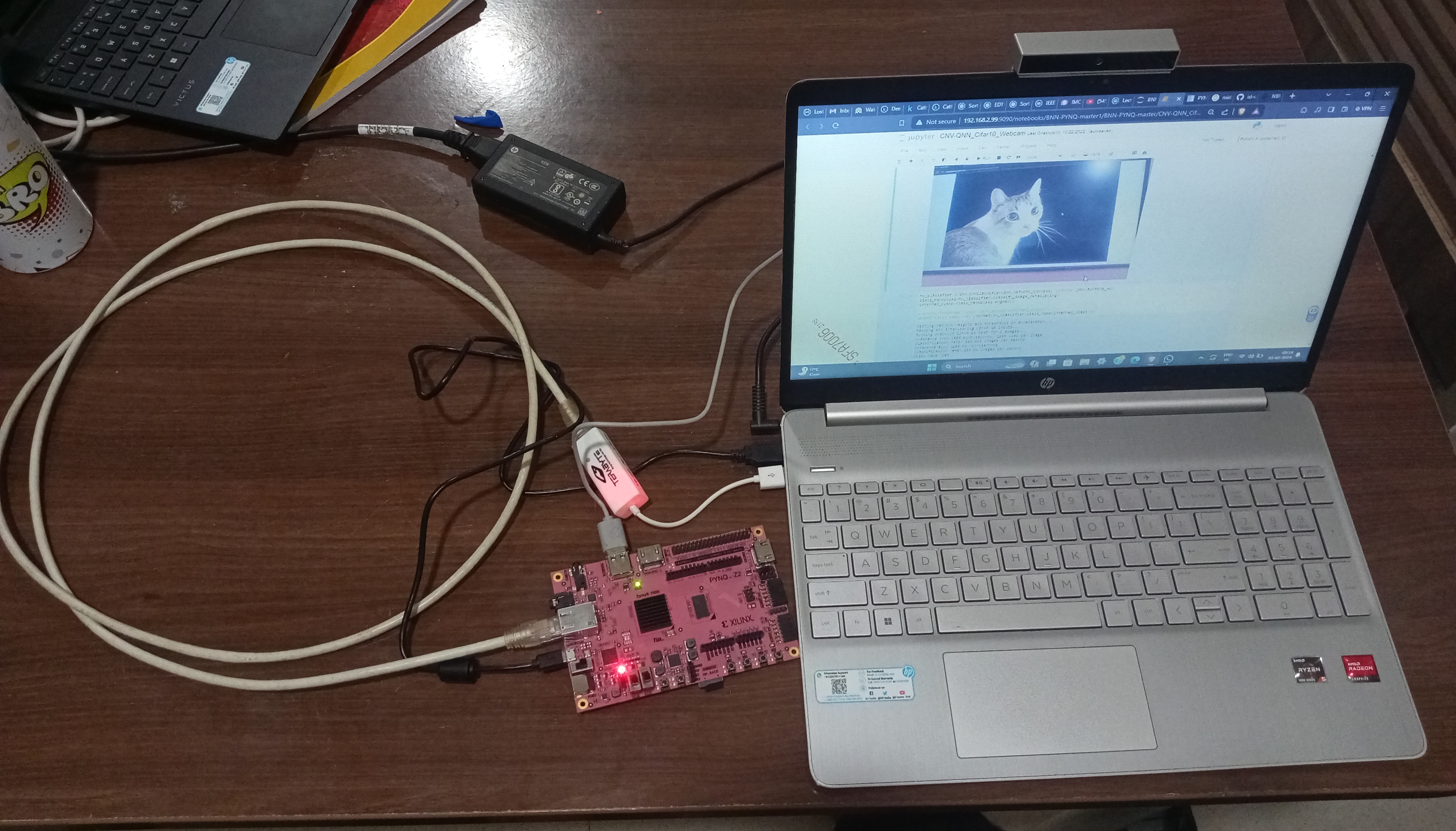}
	\caption{Hardware implementation setup for demonstration of image classification on PYNQ Z2 FPGA board.}
	\label{FPGA}
\end{figure}

\begin{figure}[t]
	\centering
	\includegraphics[scale=0.35]{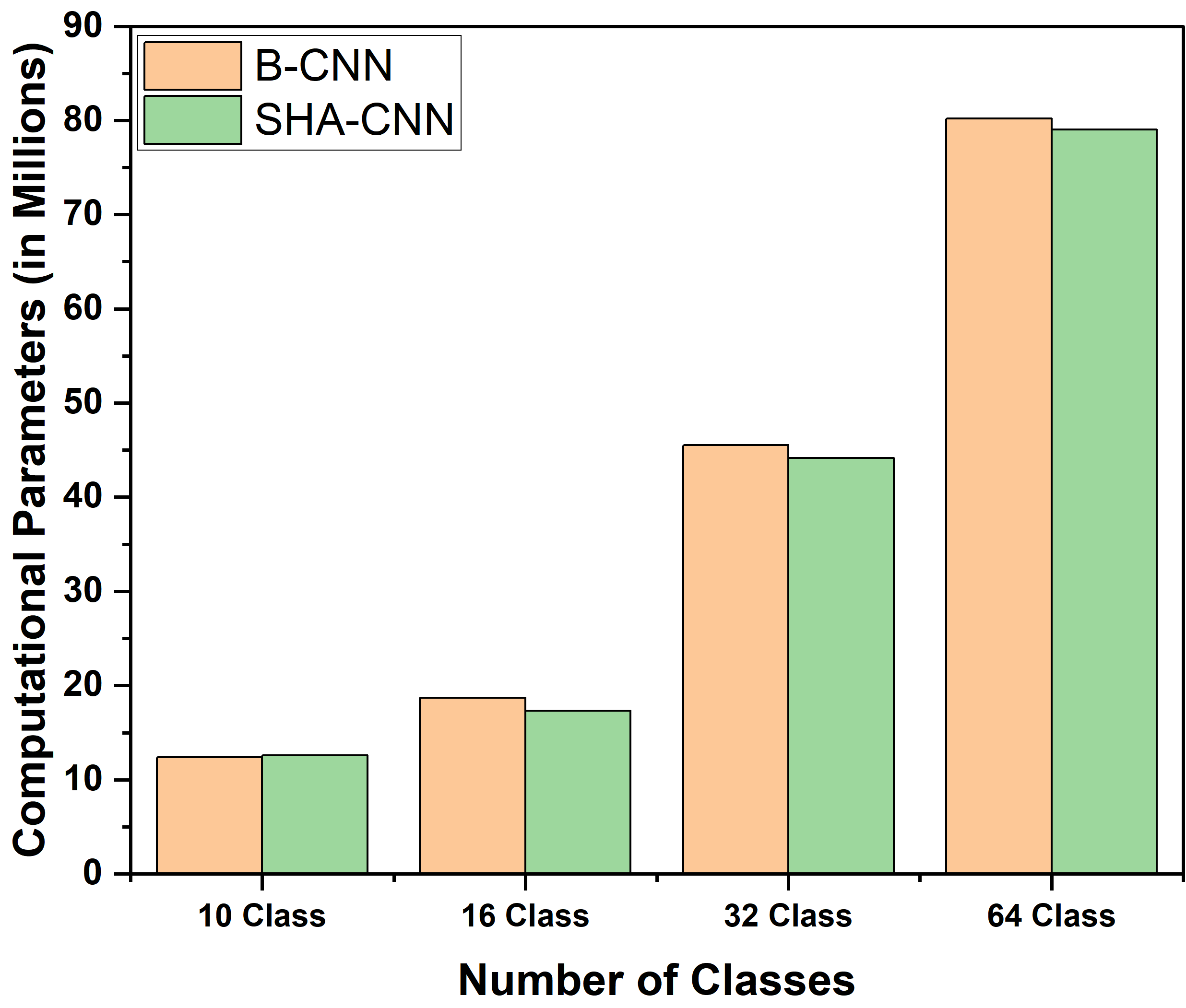}
	\caption{Computational parameters comparison with increased class of CIFAR-10.}
	\label{computations}
\end{figure}

\begin{figure}[t]
	\centering
	\subfigure[Overall MAC count]{\includegraphics[scale=0.35]{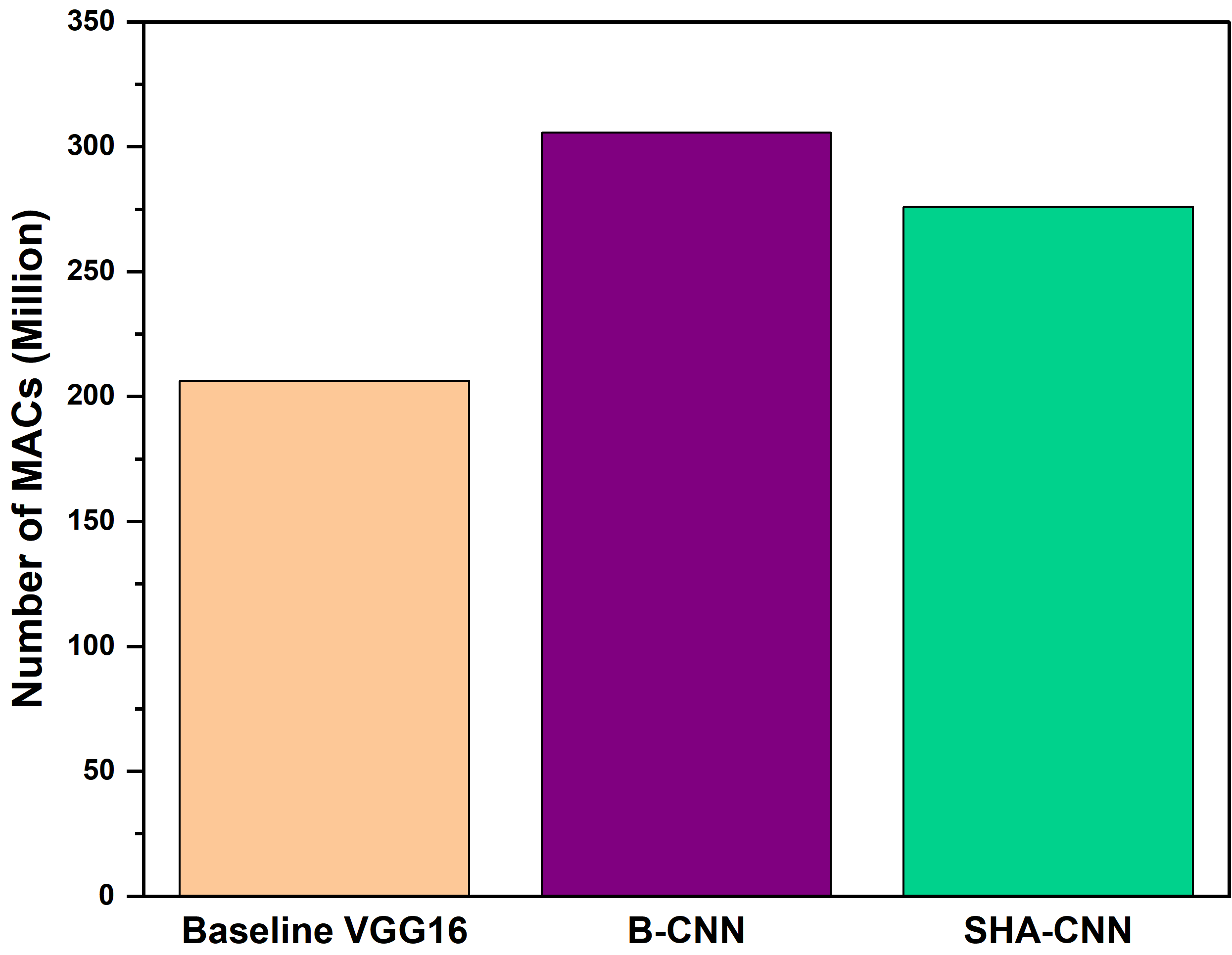}}
    \subfigure[Hierarchy level wise MAC count]{\includegraphics[scale=0.35]{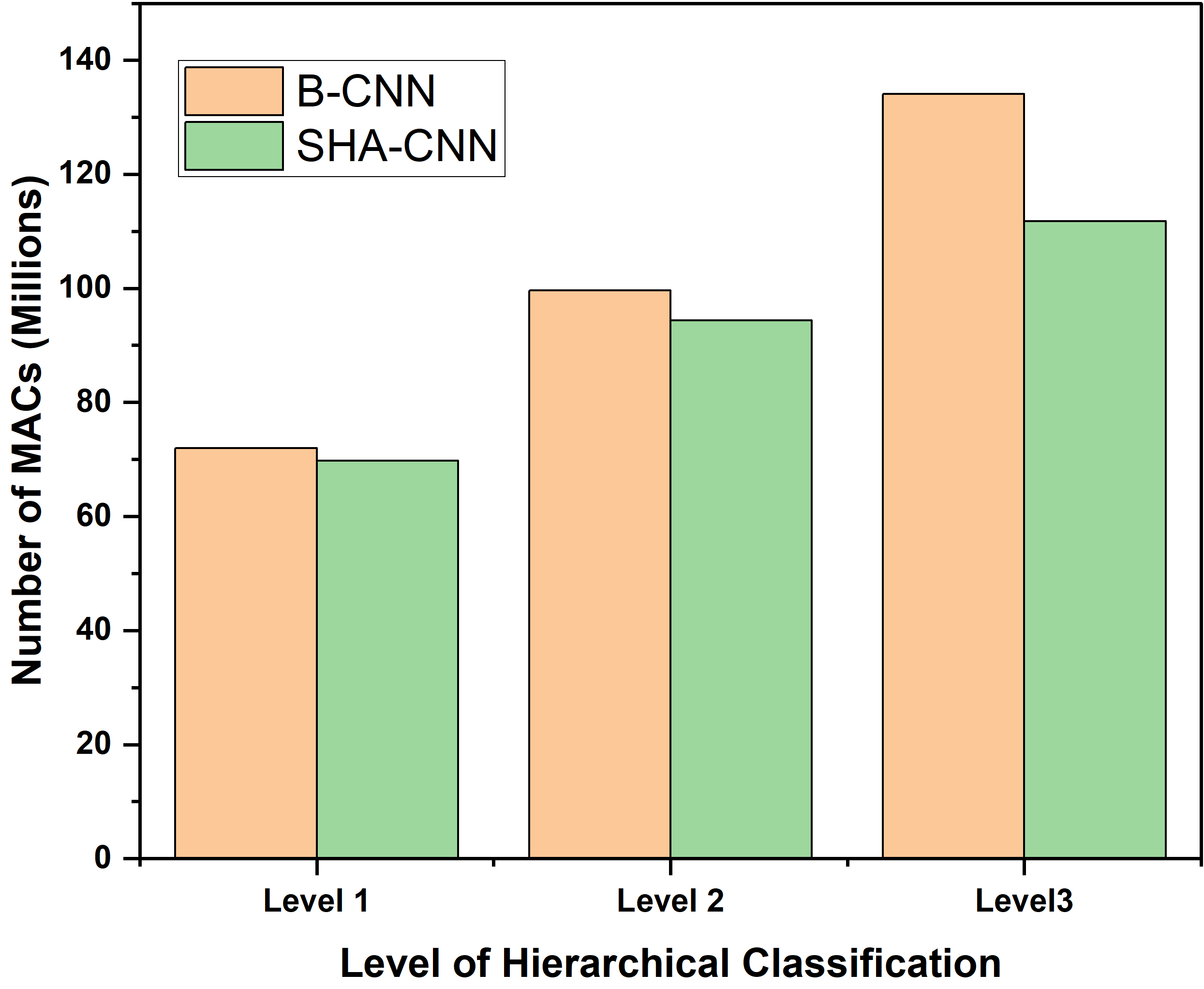}}
	\caption{Comparison of MAC computations count}
	\label{MAC}
\end{figure}

\section{Experimental Results}
Three datasets were used to experiment and validate the concepts MNIST, CIFAR-10 and CIFAR-100. The label trees for the MNIST and CIFAR-10 were taken from B-CNN \cite{BCNN} while for CIFAR-100, the label tree was created manually. The label trees for MNIST and CIFAR-10 are shown in Fig. \ref{MNISTCifar}. Test Set accuracy and catastrophic distance were used to compare model performances as shown in Table \ref{comparison}. The accuracy mentioned here is the top-1 accuracy of the model during inference. Catastrophic distance is the graphical distance between the true label and the predicted label of the data sample in the label tree as shown in Fig. \ref{cat_distance}. Graphical distance between the true and predicted class is calculated for each data sample in the test set and the average distance is reported. Fig. \ref{images} shows some sample images and their prediction at different levels. All the models were trained using Google Colab, and Python and Keras were used to write the codes. The main aim of this experiment is not to outperform the state-of-the-art models but to achieve scalability at inference, keeping in mind the applications of these models in edge devices. To achieve scalability, a compromise has to be made with accuracy. However, the proposed model performance is still comparable with state-of-the-art models while being the most scalable among all with respect to computational cost. The hardware implementation for the proposed model is performed using a PYNQ Z2 board, a camera is interfaced with the board, and the model is deployed on the board as shown in Fig. \ref{FPGA}. Fig. \ref{computations} compares the computational parameters count for B-CNN \cite{BCNN} and our proposed SHA-CNN model architecture. Here, it can be observed that as the number of classes increases, our proposed architecture uses fewer computations to achieve comparable accuracy. Fig. \ref{MAC}(a) shows the overall MAC count for the baseline VGG16 model, B-CNN model, and our proposed SHA-CNN model science. For hierarchical classification (B-CNN and SHA-CNN), the MAC counts are higher than the baseline; however, the proposed SHA-CNN has 10\% fewer MACs than B-CNN, while the accuracy of our proposed architecture is compromised with only 0.74\% for CIFAR-100 dataset. The number it requires fewer resources when deployed on edge devices. Fig. \ref{MAC}(b) shows the MACs count for level wise classification of CIFAR-100 dataset. Here we have classified the data into three level. In the first level it has 8 classes, in second level it has 20 classes and at the third level it has 100 classes. As can be seen from the Fig. \ref{MAC}(a) for each level of classification the MACs count is less for our proposed model as compared to the B-CNN. As the hierarchical level increases the MACs count decreases significantly.

\section{Conclusion}
This paper introduces a Scalable Hierarchical Aware Convolutional Neural Network (SHA-CNN) tailored for Edge AI applications. The model strikes a delicate balance between computational efficiency and accuracy, showcasing its superiority over baseline models and comparability with state-of-the-art hierarchical architectures. The innovation lies in the model's hierarchical awareness, enabling nuanced feature extraction at multiple abstraction levels. The hierarchical classification approach enhances the model's interpretability, allowing for a more granular understanding of complex datasets. Notably, SHA-CNN exhibits scalability, effortlessly accommodating new classes without extensive retraining. This adaptability is advantageous in dynamic edge environments where datasets evolve. Validation on the PYNQ Z2 FPGA board demonstrates the practical feasibility of deploying SHA-CNN in resource-constrained edge devices. SHA-CNN shows 99.34\%, 83.35\%, and 63.66\% accuracy for MNIST, CIFAR-10, and CIFAR-100 datasets, respectively. 
\IEEEpubidadjcol

\footnotesize{
\bibliographystyle{IEEEtran}
\bibliography{Reference}}

\end{document}